\documentclass[conference]{IEEEtran}
\IEEEoverridecommandlockouts
\usepackage{cite}
\usepackage{amsmath,amssymb,amsfonts}
\usepackage{algorithmic}
\usepackage{graphicx}
\usepackage{textcomp}
\usepackage{xcolor}
\def\BibTeX{{\rm B\kern-.05em{\sc i\kern-.025em b}\kern-.08em
    T\kern-.1667em\lower.7ex\hbox{E}\kern-.125emX}}

\usepackage{amsthm}

\begin{document}

\title{Synaptic Sampling of Neural Networks}

\author{\IEEEauthorblockN{James B. Aimone, William Severa, J. Darby Smith}
%\IEEEauthorblockA{\textit{Neural Exploration and Research Laboratory} \\
\textit{Neural Exploration and Research Laboratory}\\
Sandia National Laboratories, Albuquerque, New Mexico, USA\\
jbaimon@sandia.gov

}

\maketitle

\begin{abstract}
Probabilistic artificial neural networks offer intriguing prospects for enabling the uncertainty of artificial intelligence methods to be described explicitly in their function; however, the development of techniques that quantify uncertainty by well-understood methods such as Monte Carlo sampling has been limited by the high costs of stochastic sampling on deterministic computing hardware. Emerging computing systems that are amenable to hardware-level probabilistic computing, such as those that leverage stochastic devices, may make probabilistic neural networks more feasible in the not-too-distant future. This paper describes the scANN technique---\textit{sampling (by coinflips) artificial neural networks}---which enables neural networks to be sampled directly by treating the weights as Bernoulli coin flips. This method is natively well suited for probabilistic computing techniques that focus on tunable stochastic devices, nearly matches fully deterministic performance while also describing the uncertainty of correct and incorrect neural network outputs.  
\end{abstract}


\section{Overview}

Most modern artificial intelligence (AI) methods, such as artificial neural networks (ANNs), are designed to operate fully deterministically, even though the goal of most AI applications is to learn to approximate solutions. Like other approximation algorithms, the quantification of the uncertainty in ANNs is desirable, but this has remained a secondary consideration. In large part this is due to the ``Hardware Lottery'', in which deterministic algorithms, like ANNs, and deterministic computing systems, like graphics processing units (GPUs), mutually reinforce their respective technological dominance \cite{hooker2021hardware}. While stochasticity has proven to be important for effective training (i.e., stochastic gradient descent, Dropout, etc), the inference step of most ANNs is typically performed as a one-shot deterministic calculation. 

This deterministic operation of ANNs stands in stark contrast to biological neural computation, wherein neurons appear to be implementing probabilistic inference, with most (if not all) synapses in the brain operating stochastically \cite{misra2022probabilistic, malkin2023signatures}. While the search for the ideal framework to describe the brain's algorithms remains elusive \cite{aimone2023brain}, it is safe to assume that the brain's computations are more aligned to the stated goals of the Bayesian neural network (BNN) field, wherein neural networks are trained to not simply provide the most likely answer, but rather incorporate the uncertainty of the training data in its responses. 

To date, BNNs have been an intriguing area of research, but have been computationally challenging to pursue directly in part due to the high costs of sampling and the general ill fit of random number generation to GPUs, the standard workhorse of ANNs. Further, BNNs require assumptions on the structure of weight priors---typically assumed to be Gaussian---which can impact performance \cite{fortuin2021bayesian}. In this paper, we step back and consider an alternative approach; if we were to have hardware that enables inexpensive brain-like synaptic stochasticity, can we leverage that ubiquitous stochasticity to directly provide estimates of uncertainty in standard ANNs? Stated differently, does it make sense to synaptically sample trained neural networks?

Sampling neural networks is common during training with methods such as with Dropout and is effective at providing a robust training that mitigates the impact of overfitting \cite{srivastava2014Dropout}. Dropout is an example of neural sampling whereby neurons are randomly removed in training, pushing learning in each epoch to a different subset of neurons. There is also a process of synapse Dropout, which is implicit in neuron Dropout (all neurons removed by definition have their synapses removed for those training epochs), but it would be more finely administered.

It has been considered that there would be considerable value to efficiently sampling neural networks during inference. The value of sampling during inference is not one of regularization, but rather one of assessing the uncertainty inherent in a network through a sampling process. One can simply implement Dropout, as used in training, during inference \cite{gal2016Dropout}. However, since ANN models are described by the weights, it is possible that a process similar to synapse Dropout may be more suitable for inference mode. Indeed, it has been shown that weight sampling using Bernoulli Dropout or Gaussian Dropout provides a more accurate representation of uncertainty \cite{mcclure2016robustly}. Another approach envisioning probabilistic neuromorphic hardware trained neural networks with the probabilistic presence of weights as part of the model \cite{neftci2016stochastic}, illustrating that synaptic stochasticity can help realize the value of probabilistic neural algorithms like Boltzmann machines and has since been explored in with probabilistic ferroelectric synapses \cite{dutta2022neural}.

While these results are promising, one of the major challenges with sampling techniques such as Monte Carlo in assessing the uncertainty in ANNs and BNNs comes down to the computational cost \cite{papamarkou2022challenges}. Modern ANNs can often push computational limitations for one-shot deterministic inference, and repeated perturbations to the model parameters at inference time is not necessarily well-suited for computational architectures. 

Here, we examine a related method of synaptic sampling of neural networks that is related to the Bernoulli Dropout approach in \cite{mcclure2016robustly} and \cite{neftci2016stochastic}, but accounts for a hardware-specific acceleration of the restrictive sampling calculation. To motivate our approach, we will consider a Monte Carlo sampling algorithm tailored to a hypothetical stochastic neural network accelerator which uses tuned ``coinflip'' devices to represent the weights as Bernoulli probabilities. Representing the trained weights as probabilities enables direct mapping of conventionally trained ANNs into the sampling framework. By leveraging standard ANN training with only minor restrictions allows this method to be applicable to most ANN approaches, which is also a goal of spiking neural network approaches that convert ANNs to SNNs after \cite{eshraghian2023training} or at the end \cite{severa2019training} of training, rather than requiring an altogether distinct training process. Furthermore, this approach makes each sampled inference cycle extremely efficient from a computational perspective, with the effective weight matrices used only consisting of binary $0$ or $1$ values.

In this paper, we show that representing a conventionally trained ANN in this manner preserves classification accuracy in both feed-forward ANNs and convolutional ANNs in which we sample the feed-forward layers. We further illustrate how these networks provide an estimation of epistemic uncertainty of the network. Furthermore, we provide a forecast of how the representation of networks in this form enables a computationally efficient sampling strategy if appropriate hardware is used.

\section{Methods}

\subsection{Sampling (by coinflips) Artificial Neural Networks}
Each layer of a simple feed-forward neural network effectively has the following form

\begin{equation}
	 x_{i+1} = \sigma_{i+1}(W_{i} x_i + b_{i+1})
	 \label{eqn: nn_layer}
\end{equation}

where $x_{i}$ is the activation of neurons in layer $i$, $W_{i}$ is the weight matrix (i.e., synaptic strengths) between $i$ and $i+1$, $b_{i+1}$ is a bias term on each of nodes, and $\sigma()$ is the activation function of neurons, such as a rectified linear unit (ReLU) or sigmoid. We consider a specific element of $W_{i}$ as $w_{ba}$, which is the weight of the connection between node $a$ of $i$ to node $b$ of $i+1$. We consider the output of an $n$ layer neural network as $y$.

\begin{equation}
	 y = x_{n} 
	 \label{eqn: nn_layer_output}
\end{equation}

Training of our networks will be performed like typical ANN training, whereby a data set $D$ consisting of training data $[x_{\text{input}}, y_{\text{output}}]$ can be used to minimize loss between $y$ and $y_{\text{output}}$.

The vector matrix multiplication within the function is a linear mapping, and while there are some constraints on $\sigma_{i}$, in general the main requirement for ANNs is that it be non-linear and (mostly) differentiable. For most ANNs, the individual element weights of $W$ can be any real number, positive or negative, though restricted precision networks begin to introduce constraints in the dynamic range of the weights \cite{sun2020ultra}. 

\begin{figure}
	\centering
	\includegraphics[width=3.3in]{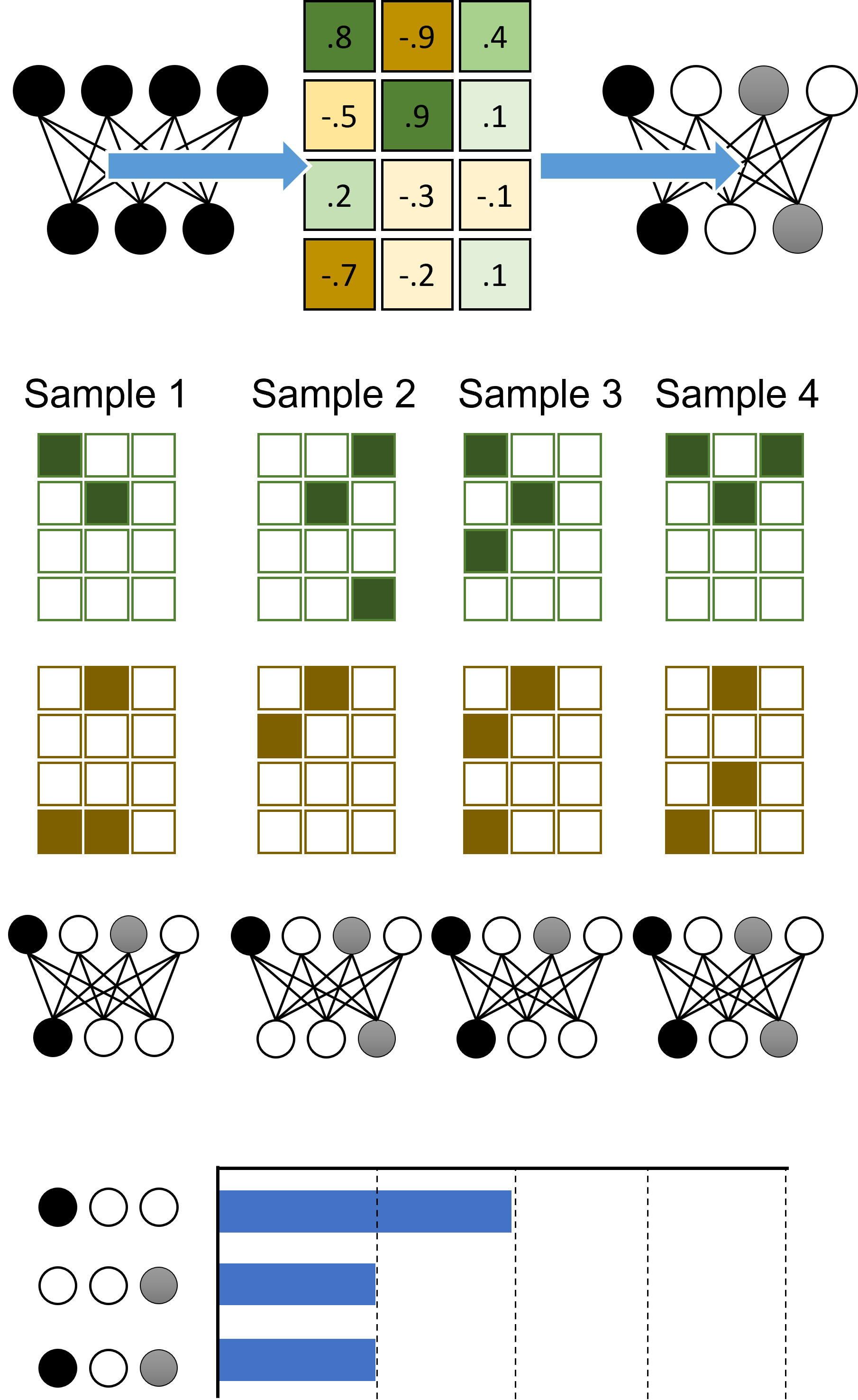}

	\caption{Illustration of scANN process. Top: Networks are trained to have weights between $-1$ and $1$. Standard deterministic operation will provide a single answer. Middle: scANN will split weight matrix into positive and negative samples, with probabilities determined by the original network weight. These sample networks will each produce an independent sampled result. Bottom: The distribution of results across samples provides a measure of the confidence of the network output.}
	\label{scANN_Overview}
\end{figure}

The approach we take here for synapse sampling is to consider $W_{i}$ as a matrix of \textit{probabilities} instead of continuous valued weights. This requires that we bound the weights in a manner that allows them to be interpreted as probabilities and to treat positive and negative weights separately. To do this, we make three modifications of a standard neural network process which are summarized in Figure \ref{scANN_Overview}.

\begin{enumerate}
	\item
	Either restrict weights in training to be between $-1$ and $1$, or renormalize the network (through modifying $b$ and $\sigma$) to allow weights to be between $-1$ and $1$. This should not have a dramatic impact on performance for standard activation functions if weights are allowed arbitrary precision.
	\item
	After training, 
	%we split the the weight matrix $W_{i}$ into positive and negative components, and interpret them as a probability matrix $P_{i}$, whereby the probability of each element being equal to $1$ is $p_{ba}$. W
	we define a random variable matrix $\Omega_i$ such that each entry $\Omega_i(b,a) \sim \text{Bernoulli}(\mathbb{P}(1) = W_i(b,a))$. %For each inference sample, $k$, instead of using $W_i$, we will instead use a sampled weight matrix $\widetilde{W}_{i}^{(k)}$, where each weight element $\widetilde{w}_{ba}$ is equal to either $1$ or $0$, with stochastic draw through a \textit{weighted coinflip} with probability $p_{ba}$ of equaling $1$. 
	%(Note: 
	As negative probabilities are not possible, we conceptually split the weight matrix, $W_i$ into positive and negative components, generating positive ($\Omega^{p}_{i}$) and negative ($\Omega^{n}_{i}$) random variables. 
	% for simplicity---and in implementation---we combine the signed components into a single sampled weight matrix, with $\widetilde{W}_{i}^{(k)}=\widetilde{W}^{p(k)}_{i} - \widetilde{W}^{n(k)}_{i}$)
	\item
	During inference, we then run $K$ sampling inference cycles, with each sample, $k$, having a \textbf{different} sample of the ($\Omega^{p}_{i}$) and negative ($\Omega^{n}_{i}$) random variables. These samples are $\widetilde{W}^{p(k)}_{i}$ and $\widetilde{W}^{n(k)}_{i}$, respectively. Each sample thus provide a set of predictions based only on using those weights that were included in that particular sample. For simplicity---and in implementation---we combine the signed components into a single sampled weight matrix, with $\widetilde{W}_{i}^{(k)}=\widetilde{W}^{p(k)}_{i} - \widetilde{W}^{n(k)}_{i}$).
\end{enumerate}	

Because the vector matrix multiplication within each neuron's input is linear, the expected value of the sample, $\mathbb{E}[\Omega_{i} x_i] = W_i x_i$, even though each sample $\widetilde{W}$ is a binary matrix. Notably, because the activation functions $\sigma_{i}$ are not linear, we should \textit{not} expect that the expected activation of neurons be the same for samples as opposed to the non-sampled network; that is $\mathbb{E}[\sigma_{i+1}(\Omega_{i} x_i+ b_{i+1})] \neq \sigma_{i+1}(W_{i} x_i+ b_{i+1})$. 

Each layer of the sampled neural network is be described by this sampled equation

\begin{equation}
	 \widetilde{x}^{(k)}_{i+1} = \sigma_{i+1}(\widetilde{W}^{(k)}_{i} \widetilde{x}^{(k)}_i + b_{i+1})
	 \label{eqn: nn_layer_sample}
\end{equation}	

and the output of our network will be defined as $\widetilde{y}^{(k)} = \widetilde{x}^{(k)}_n$.

Ultimately, we do not care about the output, $\widetilde{y}$ of any particular sample, but rather we care about the aggregate statistics across all of the samples. It is useful to use probability notation here, and we thus define a random variable $Y$ as the collection of sampled outputs from our network, and $X$ as the collection of sampled layer activations from our network. 

In our experiments, for accuracy and other metrics, we define the \textit{first choice} of the network for a given input as the most common output across all of the samples for that input, with the \textit{second choice} being the next most common output. This winner-take-all voting is an area worth further subsequent evaluation.

\subsection{Experiments}

We tested the scANN sampling approach on both feed-forward ANNs and convolutional ANNs implemented in Keras \cite{chollet2018keras}. For feed-forward networks, we examined the performance of sequential Keras networks of size $784-40-10$ on MNIST \cite{lecun1998mnist} and Fashion-MNIST \cite{xiao2017fashion} data sets. For convolutional networks, we examined the performance on MNIST and Fashion-MNIST on a simple $32$($3\times 3$)$-100-10$ network, and on CIFAR-10 \cite{krizhevsky2009learning} on $32$($3\times 3$)$-64$($3\times 3$)$-128$($3\times 3$)$-256-128-10$ networks. The emphasis on small data sets and networks is motivated by the need for hardware acceleration for embedded problems and as a proof of concept. Despite this focus on small proof-of-concept, we expect that the mechanisms described here should translate to larger-scale models.

The scANN process was implemented by initializing the input weights to the feed-forward layers to between $-0.99$ and $0.99$ and using a kernel constraint between $-1.0$ and $1.0$. Aside from the weight constraint, all training was performed conventionally with an RMSProp optimizer and categorical crossentropy loss for 100 epochs with a batch size of $100$ with Dropout. 

Regular inference was performed as a baseline control. At inference, three scANN scenarios were run for each network. First, the test data set was sampled $1000$ times with full precision weights, wherein each feed-forward input weight was treated like a probability as described above. The outputs of these inference runs were aggregated to provide a probability distribution for each test data point. 

Next, we tested lower precision inference, wherein the weights were sampled with some degree of precision noise as we would expect from stochastic devices. This lower precision sampling was implemented by rounding the Bernoulli probability of the synapse to the nearest $8$-bit (out of 256) or $4$-bit (out of 16) level. As with the full precision case, we sampled each $1000$ times and aggregated the probabilities of the outputs.

\subsection{Shannon Entropy to Estimate Uncertainty}
To measure the uncertainty represented in the sampling of the scANN networks, we used Shannon entropy to quantify the overall uncertainty in a network's sampled classification. 

\begin{equation}
	H_{item} = - \sum_{c \in class} Y_{item}(c) log_2(Y_{item}(c))
	\label{eqn: entropy}
\end{equation}

where $Y_{item}$ is the distribution of output classifications for a particular test item. The entropy measure takes no account of the `correct' classification, but rather looks only at the uncertainty within the distribution of classifications. Entropy is thus maximized when all classifications are equally likely to appear (a uniform distribution), and approaches zero when every sample is classified the same way. 

For a classification task, the maximum entropy will be $log_2(N_{classes})$. We thus can estimate the average entropy explained by the network for an item as 

\begin{equation}
	I_{item} = log_2(N_{classes}) - H_{item}
	\label{eqn: entropy}
\end{equation}

\subsection{Computational Costs}

While scANNs are able to leverage conventional training processes, scANNs will have a different costs at the inference stage of processing. The most direct increase in cost will be due to the requirement of sampling. While effective at quantifying uncertainty, the accuracy of Monte Carlo sampling scales at roughly a factor of $O(K^{0.5})$, where $K$ is the number of samples. This low sample efficiency of Monte Carlo means that many samples are often required to provide a suitable description of a model's uncertainty; particularly if the space being explored is large. This high number of samples, coupled with the high inference cost of ANNs in general, has made Monte Carlo sampling techniques such as this one prohibitive. 

While scANNs will thus require $K$ extra passes through the network, this added cost can potentially be offset by other aspects of the network. For instance, the calculation $x_A*\widetilde{W}_{AB}$ should be considerably cheaper than using $W$, because $\widetilde{W}$ is binary. This makes scANNs similar to binary weight neural networks and XNOR-nets \cite{courbariaux2015binaryconnect, rastegari2016xnor}, which offer considerable memory and inference compute cost savings, particularly if the hardware architecture is compatible. 

An additional cost to consider will be that modern ANN hardware technologies, such as GPUs and systolic arrays, are not well-configured for a rapid \textit{in situ} sampling (generating a series of different $\widetilde{W}$ matrices for each sample); instead each $\widetilde{W}$ would have to be generated from $W$ on at each sampling epoch conventionally, which is a very costly step. However, alternative technologies, such as the coinflips approach that uses stochastic behavior of magnetic tunnel junctions, may have the ability to provide this sampling natively \cite{misra2022probabilistic}.

\section{Results}
\subsection{MNIST and Fashion-MNIST scANN results}

\begin{figure}
	\centering
	\includegraphics[width=3.4in]{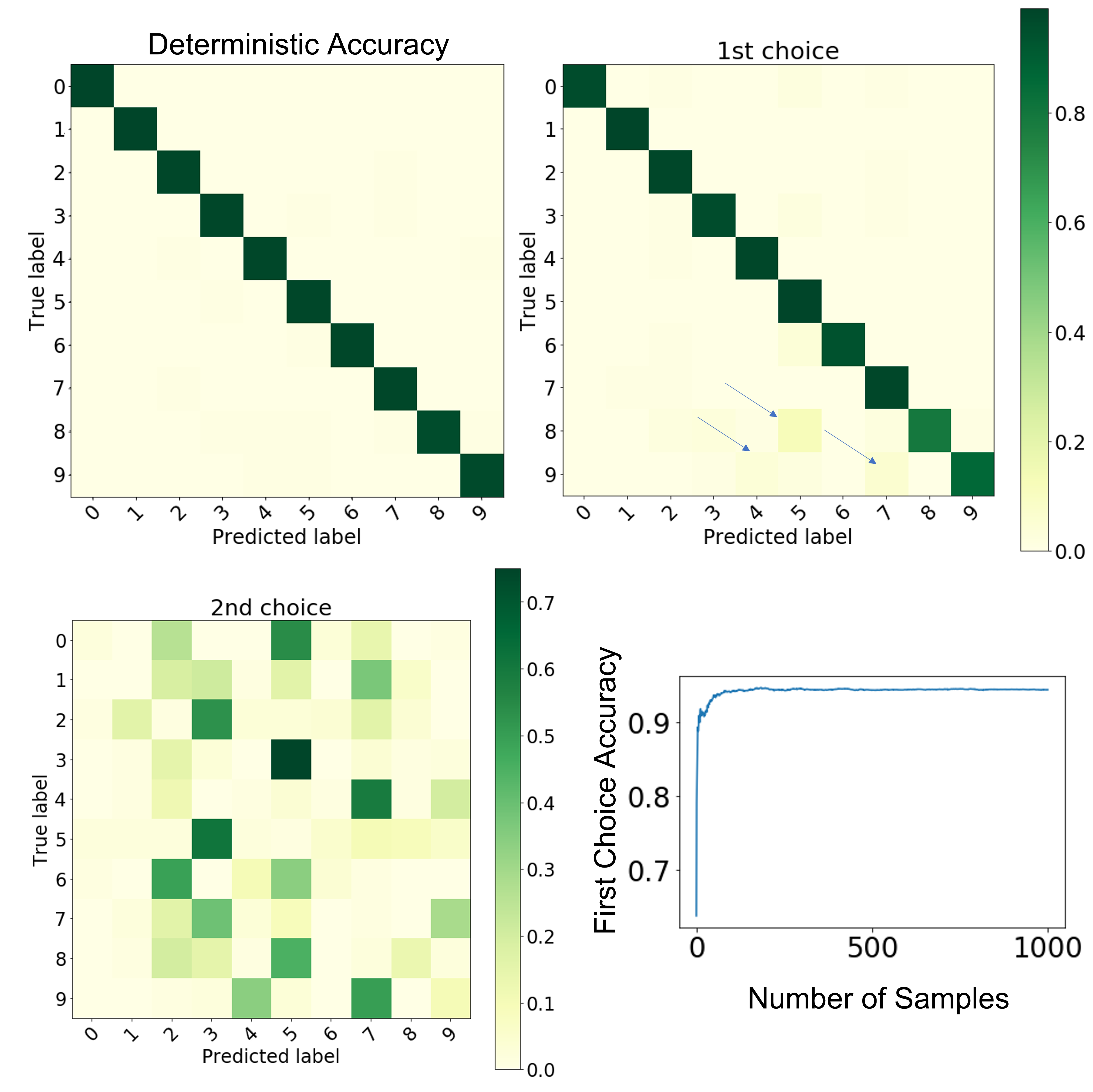}

	\caption{scANNs on MNIST. Top Left: Confusion Matrix of standard feed forward 784-400-10 network. Top Right: Confusion Matrix of first sampling choice of same network with synapse sampling. Bottom Left: Second choice of sampled network. Bottom Right: Sampling accuracy as number of samples increases.}
	\label{DetFFMNIST}
\end{figure}

\begin{table*}
\centering
 \caption{MNIST and Fashion-MNIST results}
 \label{tab:scANNResults}
 \begin{tabular}{cccccc}
  \hline
  Dataset & Network & Standard & scANN & 8-bit scANN & 4-bit scANN\\
  %& Final Energy & Subgraph \\
   & & Accuracy & Accuracy & Accuracy & Accuracy\\
  \hline
  MNIST & $784-400-10$ & 97.7\% & 94.4\% & 94.3\% & 93.7\% \\
     & $784-$[$32\times$[$3\times 3$]]$-100-10$ & 98.8\% & 98.7\% & 98.7\% & 98.6\% \\
  \hline
  FASHION- & $784-400-10$ & 87.9\% & 65.3\% & 64.2\% & 69.3\% \\
  MNIST    & $784-$[$32\times$[$3\times 3$]]$-100-10$ & 90.5\% & 90.5\% & 90.3\% & 89.2\% \\    \hline
\end{tabular}
\end{table*}

To illustrate the behavior of scANNs, we first describe the network results on simple MNIST and Fashion-MNIST feed forward and convolution networks (Table \ref{tab:scANNResults}). 

Typical MNIST results are shown in Figure \ref{DetFFMNIST}. As expected with a well trained network, the deterministic model has few mistakes that appear in a confusion matrix. The first choice of scANN networks largely match the accuracy of the deterministic network; but there are some small confusion errors (in this example, $5$ confused for $8$, $9$ confused for $4$ or $7$). Such errors are not random when one considers the nature of MNIST digits. When one considers the classifications that received the second most votes (note: this is the classification that had the second most samples that picked it first \textit{not} the next most activated output neuron), indeed one can see that the structure of MNIST digits appears quite clearly, for instance the scANN networks identifies uncertainty between $4$, $7$, and $9$ as well as uncertainty between $3$ with $2$ or $5$ (but notably not between $2$ and $5$ directly). 

Importantly, these results reflect the accuracy of the aggregate of many samples. Individually each sample network, now no longer using $W$ but rather the sampled $\widetilde{W}$, had considerably worse performance on MNIST when predicting the test set. Prediction accuracies were roughly $0.3$, with a few as low as $0.2$ and a few as high as $0.5$. To aggregate the sample performance, we combined the predictions of all of the samples. Each scANN considers that each sample of the network basically is a `vote' of the test data point's appropriate class. In MNIST, this would result in a given data point having noisy votes from each sample. For instance, a true `$4$' may get votes for each of the digits (due to vagaries of the synapse sampling), but we would expect most votes to be for $4$, while the second most votes would be for perhaps $9$, and relatively few votes for something like a $2$. 

\begin{figure*}
	\centering
	\includegraphics[width=5in]{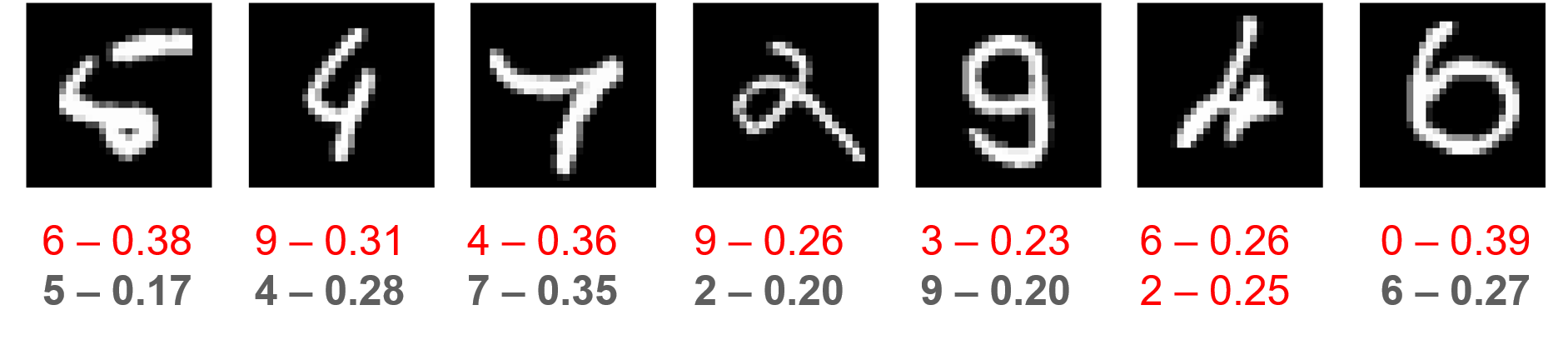}

	\caption{Examples of wrong classifications and the proportion of sample votes on example MNIST digits. Red text indicates incorrect guess and voting share; black text indicates correct guess and voting share.}
	\label{WrongExamples}
\end{figure*}

Indeed, we can look at a few specific examples of MNIST examples that are misclassified (Figure \ref{WrongExamples}). In these small networks, there are a number of difficult examples that are misclassified more often than properly classified. As can be seen by these examples, the classification that receives the second most votes is often the correct one. 

Perhaps because MNIST is not a challenging dataset for ANNs, the sample accuracy converges rather quickly with the number of samples. We observe a different outcome with the same feed forward network processing Fashion-MNIST \cite{xiao2017fashion}. Fashion-MNIST was designed with this use case in mind; providing a different and arguably more difficult data set that is structured identically to MNIST. Rather than digits, Fashion-MNIST has several clothing categories, some of which are highly similar to one another (for example, there are several types of footwear: sandals (5), sneakers (7), and boots (9))

\begin{figure}
	\centering
	\includegraphics[width=3.3in]{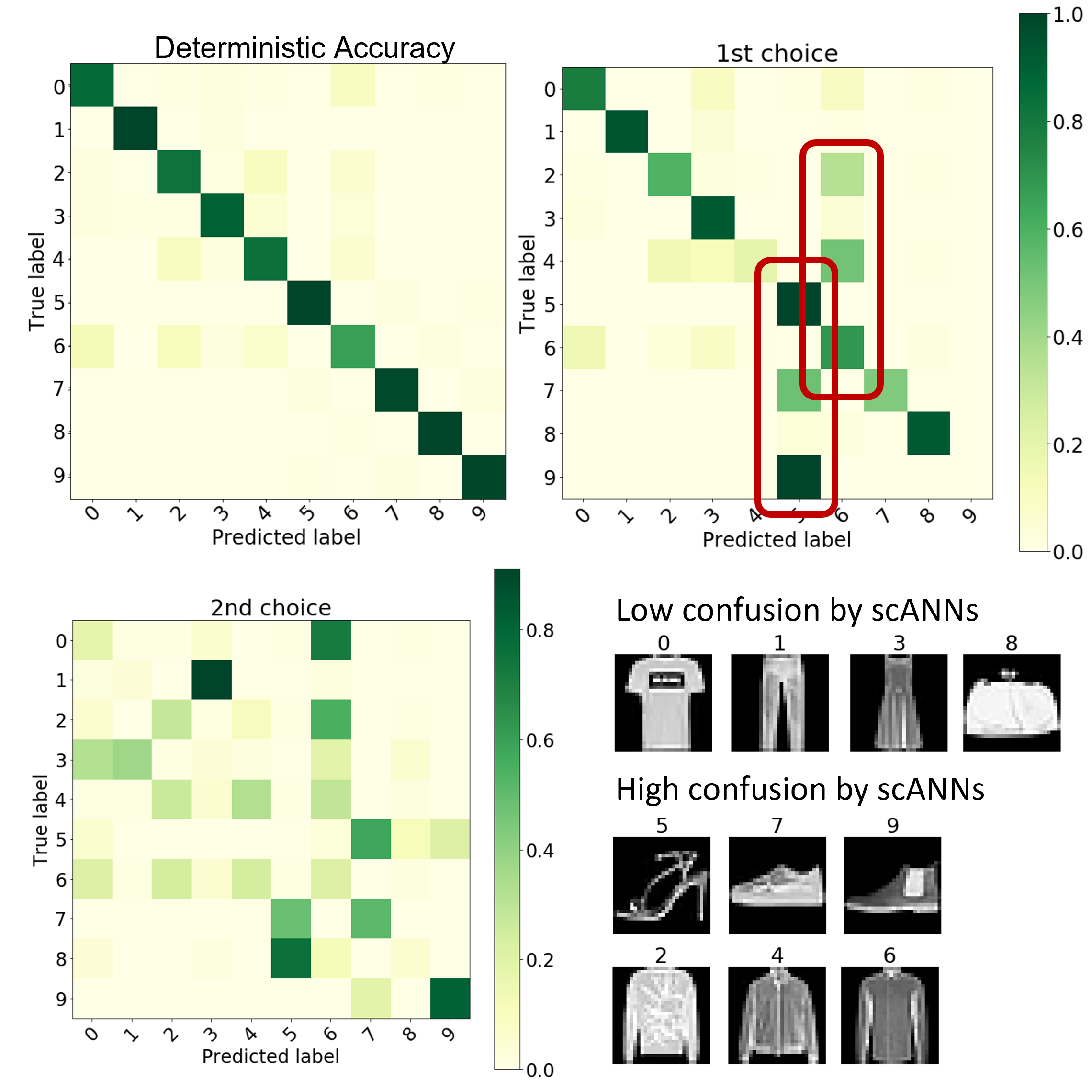}

	\caption{scANNs on Fashion. Top Left: Confusion Matrix of standard feed forward 784-400-10 network. Top Right: Confusion Matrix of first sampling choice of same network with synapse sampling. Bottom Left: Second choice of sampled network. Bottom Right: Example images from each Fashion-MNIST class, sorted into clusters scANNs observe.}
	\label{DetFFFashion}
\end{figure}

As may be predicted from the increased uncertainty inherent in the Fashion-MNIST data set, we see that the feed forward scANNs do not match the deterministic network's performance as well as we saw for standard MNIST (Figure \ref{DetFFFashion}). Rather, we see that for sampled network, the first choice of the scANNs often was confused with other examples from similar classes, for instance sneakers and boots often are highly likely to be predicted as sandals. While boots are almost always misclassified as `sandal' in the plurality of samples, the second most correct samples are in fact classified as the correct answer of `boot'. This further indicates that the scANNs are capturing the structure of this data set's higher relative uncertainty. 

Notably, the use of convolutional networks rather than feed-forward networks appears to improve the accuracy of the sampling networks. In our convolutional scANNs, we restricted the sampling to only the feed-forward layers of the networks. As in the fully feed-forward example, sampling was similarly made on one dense hidden layer (following the convolution layer) and the dense output layer. 

\begin{figure}
	\centering
	\includegraphics[width=3.7in]{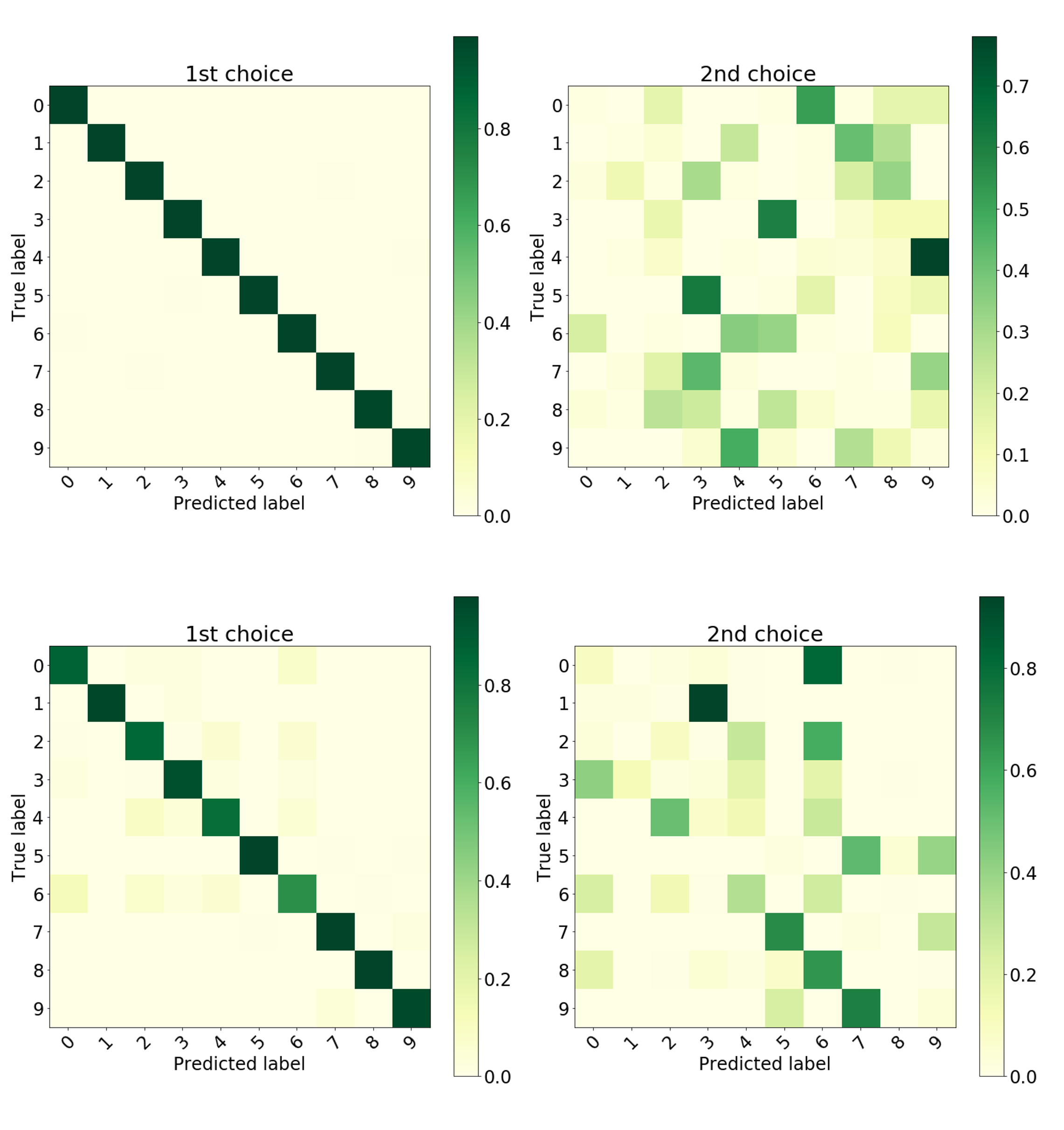}

	\caption{Performance of convolutional scANNs. Top: MNIST sampling results. Bottom: Fashion-MNIST sampling results.}
	\label{Conv_Results}
\end{figure}

Not surprisingly, the baseline accuracy of the convolutional networks was higher for both MNIST and Fashion-MNIST (Table \ref{tab:scANNResults}). Notably, however, the scANN accuracy was far more preserved in the convolutional networks; suggesting that sampling the dense layers following trained convolutional filters is more stable than sampling the raw input or that larger networks provide increased sampling accuracy. Interestingly, the second choice outputs are not identical to the second choice in the feed-forward networks (Figure \ref{Conv_Results}). 

\begin{figure*}
	\centering
	\includegraphics[width=6.2in]{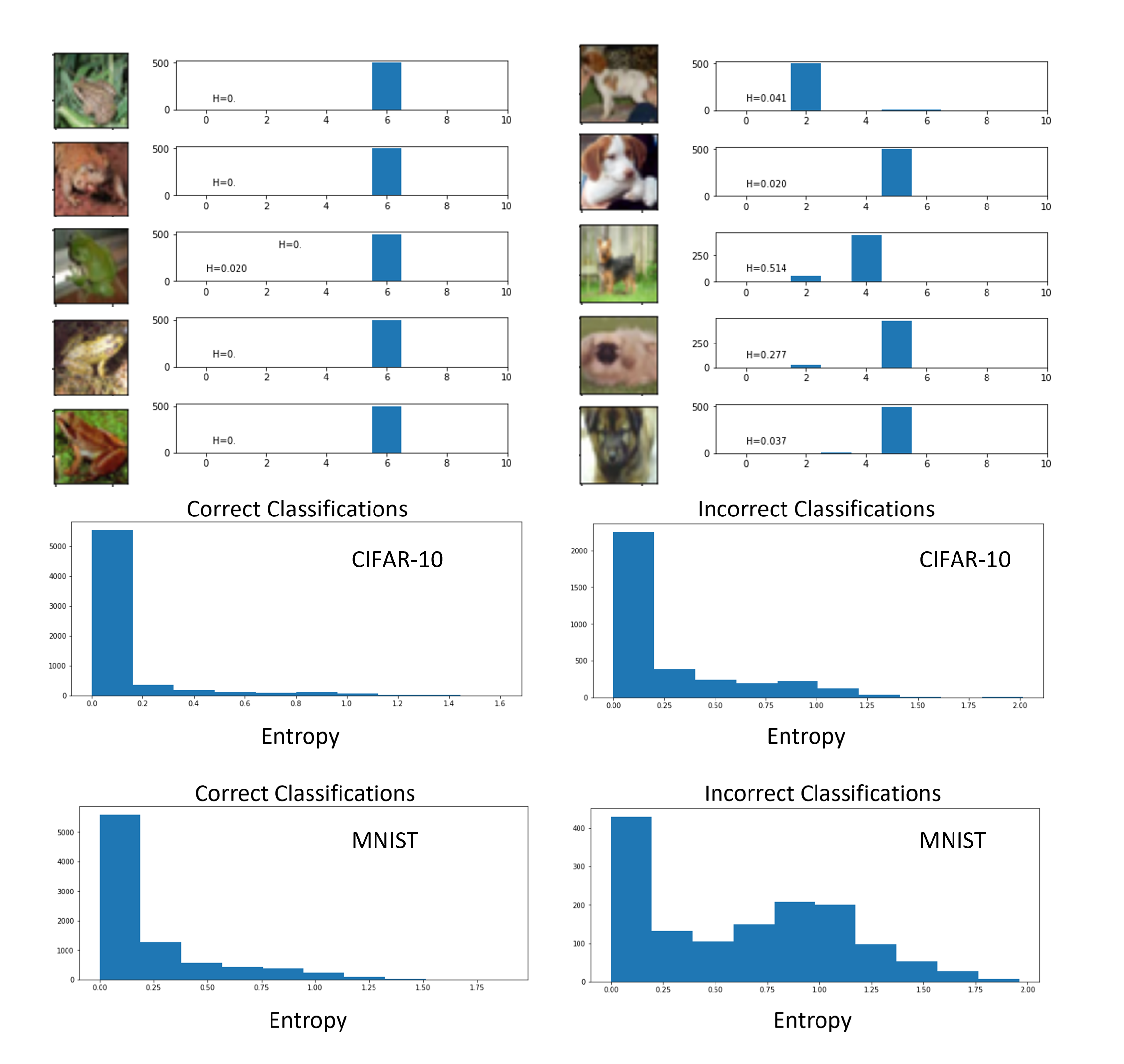}

	\caption{Entropy of sampled CIFAR networks. Top: Five illustrative examples are shown from `frog' and `dog' CIFAR classes. The scANN networks accurately classify these examples, but there is slightly higher entropy in several of the classifications. Middle: Histograms show distribution of entropy values of correct and incorrect scANN classifications for CIFAR-10 networks. Incorrect samples typically show higher entropy, as would be expected from increased uncertainty. Bottom: Distribution of entropy of MNIST networks.}
	\label{MNIST_Entropy}
\end{figure*}

\subsection{Entropy of sampling MNIST and CIFAR networks}

We can systematically evaluate the effectiveness of the sampling in characterizing the uncertainty of the data set. Irrespective of right or wrong, for each test set data point, the scANN approach provides a distribution of classifications. The distribution of these classifications is a measure of the network's confidence. If every sample provides the same output, there is minimal entropy in the distribution, or equivalently the network is communicating high information about what the output class is. However, if the samples of a data point are highly variable, there is accordingly higher entropy and lower information content (Figure \ref{MNIST_Entropy}). 

While entropy itself is not a measure of accuracy, ideally a well functioning network should be more confident in classifications in which it is correct and less confident for classifications in which it is incorrect. In our MNIST and CIFAR scenarios, we accordingly observed that those classifications in which the network was correct were largely low entropy classifications, whereas the classifications in which the plurality of votes were incorrect had a much higher probability of being high entropy.
\begin{figure*}
	\centering
	\includegraphics[width=6.4in]{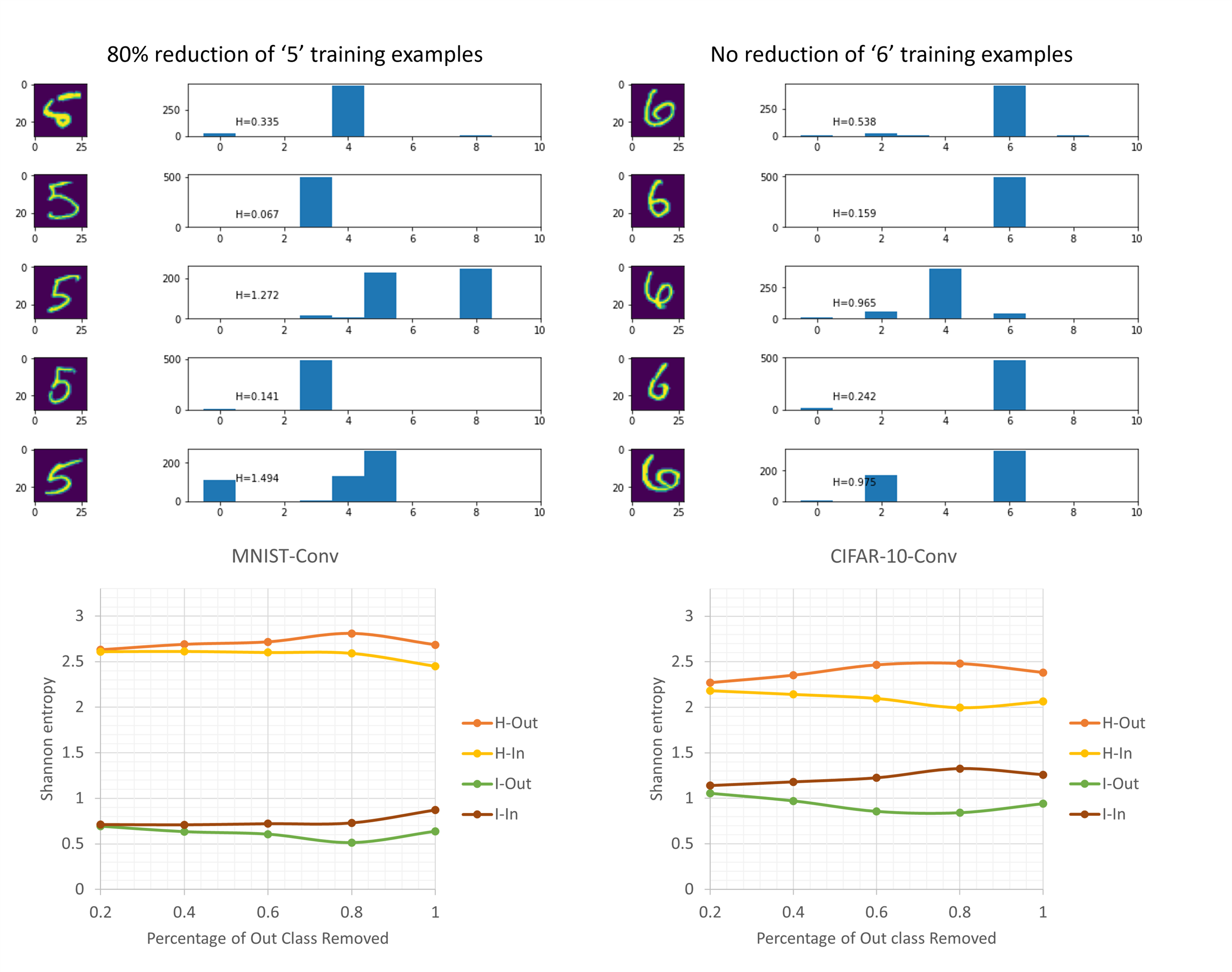}

	\caption{MNIST scANN networks with training data withheld from certain classes exhibit more uncertainty in the classifications of reduced classes. Bottom: Quantification of entropy and information of MNIST and CIFAR-10 networks at varying levels of held-out training classes. H-Out and I-Out refers to entropy and information of withheld class, H-In and I-In refers to entropy of normally trained classes.}
	\label{MNIST_Entropy_5}
\end{figure*}

\subsection{Increased Uncertainty in Underrepresented Classes}
A challenge in evaluating the contribution of sampling a neural network is that there is generally is not a fundamental ground truth of a data set's uncertainty. However, one option is that we can modify training sets to under represent certain classes in training, which we would expect to increase the uncertainty of those items relative to those classes that are represented fully. 

To accomplish this, we modified both CIFAR-10 and MNIST to partially eliminate a fraction of the training sets for one class. We then computed the average Shannon entropy (Equation \ref{eqn: entropy}) for items in the underrepresented class compared to items that are fully represented. We expect that there will be two implications of this hold-out experiment: in addition to the naive prediction that performance of networks will decrease when a particular class is underrepresented, we also hypothesize that the entropy associated with the impacted class will preferentially increase---in effect, the sampling will reveal a broader diversity of answers for a relatively weakly trained class relative to the control classes. 

Indeed, we observe exactly that in aggregate when networks trained on biased training sets are sampled. When networks are trained on MNIST with one class underrepresented, the average entropy of the withheld class increases and the associated information about those data points decreases. Note that there is no explicit information that the network is misclassifying the digits from withheld classes, but the network shows generally higher entropy, especially with the underrepresented class (Figure \ref{MNIST_Entropy_5}A). Notably, these results are comparable in CIFAR-10 data as well.

\subsection{The impact of randomness precision on sampling classification}
By controlling randomness at the device level through the use of Bernoulli coin flips, we believe that we can make stochasticity a universal resource \cite{misra2022probabilistic}. However, while we consider here what we can do with a large number of weighted coins, we must also consider that the weighting of that coin likely cannot be arbitrarily set. Rather, the weight will have some precision, and that precision will undoubtedly be lower than the floating point precision typically used in ANN weights.

We then repeated the same sampling process, but instead of comparing the weights to a computer-precision uniform random number, we instead compared to a random number at either $8$-bit or $4$-bit precision. In our simulations, restricting the precision of scANN networks has a small, but not substantial, effect on sampling accuracy (Table \ref{tab:scANNResults}) in both MNIST and Fashion-MNIST. Even considerably lower precision networks (4-bit) still maintain most of their classification ability. This suggests that if information can be moved into the sampling regime, the loss of information due to lower precision can perhaps in part be compensated for by the sampling process. 

\begin{figure}[!t]
	\centering
	\includegraphics[width=3.5in]{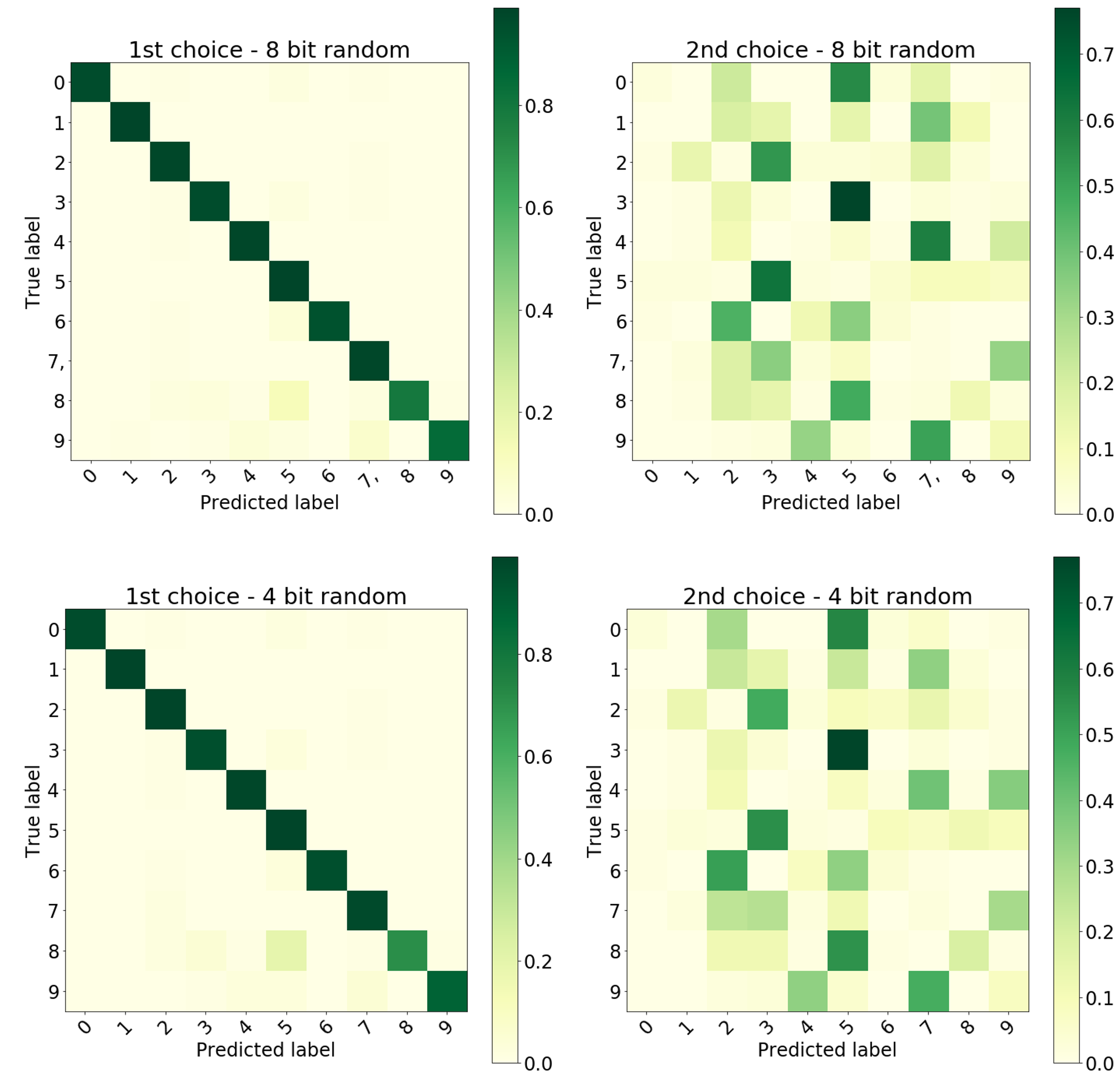}

	\caption{784-400-10 8-bit and 4-bit Sampling Approximation on MNIST.}
	\label{Combined400Plots_lowprecision}
\end{figure}

\section{Conclusion}

Combined, these results support the idea that sampling can be performed on ANNs, that the approach is not overly prohibitive (can achieve near deterministic accuracy with ~1000 samples), and appear to offer something beyond a straight deterministic solution in looking a distributions of `sample votes'. 

While the scANN approach is clearly not well suited for conventional AI computing platforms, the potential of tunable stochastic devices to implement opens up a possibility that a direct hardware instantiation of these networks would be highly efficient; possibly providing the added value of Monte Carlo sampling of ANNs for minimal cost. There is increasing work in identifying probabilistic devices such as p-bits \cite{camsari2019p} and related coinflip devices \cite{misra2022probabilistic} that are amenable to being used in networks such as this. 

Importantly, the value of device-level sampling will further require a direct neuromorphic hardware instantiation as well. A related approach illustrates how using analog crossbars with stochastic MTJs as noisy weights can implement BNNs \cite{liu2022bayesian}. Importantly, unlike this approach which matches Gaussian weight noise of a BNN to the analog noise of an MTJ, scANNs directly convert deterministic ANNs for probabilistic sampling. This ability to use standard ANN training both offers greater training flexibility and lower demands on stochastic devices, but does make the scANN method deviate from other BNN formalisms.

Because the networks described here do not leverage a Bayesian training approach, an important question is ``what do these distributions represent?'' As can be seen from the data here, the confusion matrices of second choices for both MNIST and Fashion-MNIST are indicative that there is some underlying structure that this approach is identifying. Thus, while it is not immediately obvious what a second place tally actually represents, it does suggest that there are certain relationships in these data sets that are reliably seen as second choices. That these networks extract this structure is not a trivial outcome. In effect, networks were only trained on the diagonal (the labels represent the `correct' choices) of these confusion matrices, the off-diagonal terms are all equivalently `wrong'. This confirms an intuition that standard ANNs do preserve some of the underlying structure of the relationships between classes; and the scANN approach appears capable of extracting it. 

Finally, the lower precision examination here provides some encouragement that the approach is amenable for hardware acceleration if ubiquitous random numbers can be generated. We and others have demonstrated that tunable coinflip MTJ devices are reasonable \cite{rehm2023stochastic, liu2022random}, but almost certainly the precision of these numbers will not be particularly high---especially in systems at large scale. While the 8-bit and 4-bit studies had modestly increased error, both were run with relatively few samples. Further work will be necessary to determine if this robustness to low precision is due to this probabilistic formulation being inherently more precision tolerant or is related to known benefits of stochastic rounding in low-precision systems. Regardless, the distributions of second choices in the reduced-precision networks still had the rough structure as the full precision network, indicating that the value of sampling to describe uncertainty are preserved in low-precision conditions. This is encouraging and suggests that this approach may provide utility in scalable probabilistic systems.

\section*{Acknowledgements}
We thank Lindsey Aimone for proof reading this manuscript. This research was funded by the Department of Energy Office of Science (ASCR/BES) Microelectronics Co-Design program COINFLIPS. 
This article has been authored by an employee of National Technology \& Engineering Solutions of Sandia, LLC under Contract No. DE-NA0003525 with the U.S. Department of Energy (DOE). The employee owns all right, title and interest in and to the article and is solely responsible for its contents. The United States Government retains and the publisher, by accepting the article for publication, acknowledges that the United States Government retains a non-exclusive, paid-up, irrevocable, world-wide license to publish or reproduce the published form of this article or allow others to do so, for United States Government purposes. The DOE will provide public access to these results of federally sponsored research in accordance with the DOE Public Access Plan https://www.energy.gov/downloads/doe-public-access-plan.

This paper describes objective technical results and analysis. Any subjective views or opinions that might be expressed in the paper do not necessarily represent the views of the U.S. Department of Energy or the United States Government.

\bibliographystyle{plain}
%\bibliography{scann}

\end{document}